%% file: eacl2021.tex
\pdfoutput=1
\documentclass[11pt]{article}

\usepackage{acl}

\usepackage{times}
\usepackage{latexsym}
\usepackage{comment}
\usepackage{url}

\usepackage{tabularx}
\usepackage{adjustbox}
\usepackage{rotating}
\usepackage{xcolor}
\usepackage{color, colortbl}
\usepackage{multirow}
\usepackage{tablefootnote}
\usepackage{comment}
\usepackage{float}
\usepackage{refcount}
\usepackage{subcaption}
\usepackage{paralist}
\usepackage{amsfonts}
\usepackage{amssymb}
\usepackage{booktabs}
\usepackage{url}
\usepackage[multiple]{footmisc}
\usepackage{hyperref}
\usepackage{microtype}
\usepackage{tablefootnote}
\usepackage{footnote}
\makesavenoteenv{tabular}

\newcommand{\tabincell}[2]{\begin{tabular}
{@{}#1@{}}#2\end{tabular}}

\title{Meta Learning for Natural Language Processing: A Survey}

\author{Hung-yi Lee \\
National Taiwan University\\
  \texttt{\small hungyilee@ntu.edu.tw} \\\And
Shang-Wen Li \\
Amazon AI\thanks{$^*$Work done while working at Amazon Inc. The current affiliation is Meta AI.}\\
  \texttt{\small shangwel@amazon.com} \\ \And
Ngoc Thang Vu \\
University of Stuttgart\\
  \texttt{\small thangvu@ims.uni-stuttgart.de}
  }

\begin{document}

\maketitle
\begin{abstract}
Deep learning has been the mainstream technique in natural language processing (NLP) area. However, the techniques require many labeled data and are less generalizable across domains. Meta-learning is an arising field in machine learning studying approaches to learn better learning algorithms. Approaches aim at improving algorithms in various aspects, including data efficiency and generalizability. Efficacy of approaches has been shown in many NLP tasks, but there is no systematic survey of these approaches in NLP, which hinders more researchers from joining the field. Our goal with this survey paper is to offer researchers pointers to relevant meta-learning works in NLP and attract more attention from the NLP community to drive future innovation. This paper first introduces the general concepts of meta-learning and the common approaches. Then we summarize task construction settings and application of meta-learning for various NLP problems and review the development of meta-learning in NLP community.

%such as text classification, machine translation, dialog state tracking and knowledge base completion. 

\end{abstract}

\section{Introduction} % (Daniel)
Recently, deep learning (DL) based natural language processing (NLP) has been one of the research mainstreams and yields significant performance improvement in many NLP problems. However, DL models are data-hungry. The downside limits such models' application to different domains, languages, countries, or styles because collecting in-genre data for model training are costly. 
%The diversity in human language problems makes challenges even more significant. 

To address the challenges, meta-learning techniques are gaining attention. 
Meta-learning, or Learning to Learn, aims to learn better learning algorithms, including better parameter initialization~\cite{Finn:ICML17}, optimization strategy~\cite{Andrychowicz:NIPS16,Ravi:ICLR17}, network architecture~\cite{NASICLR17,NASCVPR18,NASICML18}, distance metrics~\cite{Vinyals:NIPS16,Gao:AAAI19,Sung:CVPR18}, and beyond~\cite{Mishra:ICLR18}. 
Meta-learning allows faster fine-tuning, converges to better performance, yields more generalizable models, and it achieves outstanding results for few-shot image classificaition~\cite{Triantafillou2020Meta-Dataset}. %OMNIGLOT, miniimage net, Meta world, Meta dataset (https://arxiv.org/abs/1903.03096) %~\cite{EffectiveFinetuneNeurIPS2021}  ~\cite{TaskRobustNeurIPS2020,} ~\cite{Triantafillou2020Meta-Dataset}
The benefits alleviate the dependency of learning algorithms on labels and make model development more scalable.
%\textcolor{black}{For the general concept of meta-learning and its performance in image processing, please refer to the overview papers for general meta-learning technology~\cite{MetaSurveyHuisman,MetaSurveyTimothy}.}
Image processing is one of the machine learning areas with abundant applications and established most of the examples in the previous survey papers on  meta-learning~\cite{MetaSurveyTimothy,MetaSurveyHuisman}.
%\footnote{Tutorial in ICML 2019: \url{https://sites.google.com/view/icml19metalearning}}. 
%\footnote{Course at Stanford: \url{http://cs330.stanford.edu/}}.

On the other hand, there are works showing benefits of meta-learning techniques in performance and data efficiency via applying meta-learning to NLP problems. %from relation extraction and machine translation to dialogue generation and state tracking.
Please refer to Tables~\ref{table:methods} and~\ref{table:methods2} in the appendix for NLP applications improved by meta-learning. 
Tutorial~\cite{lee2021meta} and Workshop~\cite{lee2021proceedings} are organized at ACL 2021 to encourage exchange and collaboration among NLP researchers interested in these techniques.
%However, to the best of our knowledge, there is no survey paper on the applications of meta-learning to NLP.
To facilitate more NLP researchers and practitioners benefiting from the advance of meta-learning and participating in the area, we provide a systematic survey of meta-learning to NLP problems in this paper. 
\textcolor{black}{There is another survey paper on meta-learning in NLP~\cite{YinSurvey}. 
While~\citet{YinSurvey} describes meta-learning methods in general, this paper focuses on the idea of making meta-learning successful when applied to NLP and provides a broader review of publications on NLP meta-learning. 
This paper is organized as below.
\begin{itemize}
   \item A brief introduction of meta-learning backgrounds, general concepts, and algorithms in Section~\ref{sec:background}.
  \item  Common settings for constructing meta-learning tasks in Section~\ref{sec:task}.
  \item  Adaptation of general meta-learning approaches to NLP problems in Section~\ref{sec:specific}.
  \item  Meta-learning approaches for special topics, including knowledge distillation and life-long learning for NLP applications in Section~\ref{sec:advance}. 
\end{itemize}
%This survey paper intends to provide researchers with pointers to recent meta-learning breakthroughs in the NLP community, help them understand the techniques, and inspire more innovation. 
Due to space constraints, we will not give too many detailed descriptions of general meta-learning techniques in this survey paper. For general concepts of meta-learning, we encourage readers to read the previous overview paper~\cite{YinSurvey,MetaSurveyTimothy,MetaSurveyHuisman}. }

%We demonstrate the capability of meta-learning and its impact on broad NLP problems, including but not limited to machine translation (MT), natural language inference (NLI), dialog state tracking (DST), and question answering (QA). 

\section{Background Knowledge for Meta Learning} %(Hung-yi)
\label{sec:background}

The goal of machine learning (ML) is to find a function $f_{\theta}(x)$ parametrized by model parameters $\theta$ for inference from training data. 
For machine translation (MT), the input $x$ is a sentence, while $f_{\theta}(x)$ is the translation of $x$; for automatic speech recognitoin (ASR), $x$ is an utterance, while $f_{\theta}(x)$ is the transcription; 
In DL, $\theta$ are the network parameters, or weights and biases of a network. 
To learn $\theta$, there is a loss function $l(\theta;\mathcal{D})$, where $\mathcal{D}$ is a set of paired examples for training,
\begin{equation}
\mathcal{D}=\{(x_1,y_1),(x_2,y_2),...,(x_K,y_K)\},  \label{eq:D}   
\end{equation}
where $x_k$ is function input, $y_k$ is the ground truth, and $K$ is the number of examples in $\mathcal{D}$.
The loss function $l(\theta;\mathcal{D})$ is defined as below:
\begin{equation}
    l(\theta;\mathcal{D}) = \sum_{k=1}^K d(f_{\theta}(x_k), y_k). 
    \label{eq:small_l}
\end{equation}
where $d(f_{\theta}(x_k),y_k)$ is the ``distance'' between the function output $f_{\theta}(x_k)$ and the ground truth $y_k$.
For classification problem, $d(.,.)$ can be cross-entropy; for regression, it can be L1/L2 distance. 
The following optimization problem is solved to find the optimal parameter set $\theta^*$ for inference via minimizing the loss function $l(\theta;\mathcal{D})$.
\begin{equation}
    \theta^* = \arg \min_{\theta} l(\theta;\mathcal{D}). \label{eq:optmization}
\end{equation}

In meta-learning, what we want to learn is a learning algorithm.
The learning algorithm can also be considered as a function, denoted as $F_{\phi}(.)$.
The input of $F_{\phi}(.)$ is the training data, while the output of the function $F_{\phi}(.)$ is the learned model parameters, or $\theta^*$ in (\ref{eq:optmization}).
The learning algorithm $F_{\phi}(.)$ is parameterized by \textit{meta-parameters} $\phi$, which is what we want to learn in meta-learning.
If $F_{\phi}(.)$ represents gradient descent for deep network, $\phi$ can be initial parameters, learning rate, network architecture, etc. 
Different meta-learning approaches focus on learning different components.
For example, model-agnostic meta-learning (MAML) focuses on learning initial parameters~\cite{Finn:ICML17}, which will be further descried in Section~\ref{subsec:INIT}.
Learning to Compare methods like Prototypical Network~\cite{Snell:NIPS17} in Section~\ref{subsec:COM} learn the latent representation of the inputs and their distance metrics for comparison. 
Network architecture search (NAS) in Section~\ref{subsec:NAS} learns the network architecture~\cite{NASICLR17,NASCVPR18,NASICML18}.

To learn meta-parameters $\phi$, \textit{meta-training tasks} $\mathcal{T}_{train}$ are required.  
\begin{equation}
\mathcal{T}_{train} = \{\mathcal{T}_1,\mathcal{T}_2,...,\mathcal{T}_N\}, 
\label{eq:T_train}
\end{equation}
where $\mathcal{T}_n$ is a task, and $N$ is the number of tasks in $\mathcal{T}_{train}$. 
Usually, all the tasks belong to the same NLP problem; for example, all the $\mathcal{T}_n$ are QA but from different corpora, but it is also possible that the tasks belong to various problems. %this sentence is redundant
Each task $\mathcal{T}_n$ includes a \textit{support set} $\mathcal{S}_n$ and a \textit{query set} $\mathcal{Q}_n$.
Both $\mathcal{S}_n$ and $\mathcal{Q}_n$ are paired examples as $\mathcal{D}$ in (\ref{eq:D}).
The support set plays the role of training data in typical ML, while the query set can be understood as the testing data in typical ML.
However, to not confuse the reader, we use the terms support and query sets in the context of meta-learning instead of training and testing sets. 

In meta-learning, there is a loss function $L(\phi;\mathcal{T}_{train})$, which represents how ``bad'' a learning algorihtm paramereized by $\phi$ is on  $\mathcal{T}_{train}$.
$L(\phi;\mathcal{T}_{train})$ is the performance over all the tasks in $\mathcal{T}_{train}$,
\begin{equation}
L(\phi;\mathcal{T}_{train}) = \sum_{n=1}^N l(\theta^n;\mathcal{Q}_n). \label{eq:meta-optimization}
\end{equation}
The definition of the function $l(.)$ above is the same as in (\ref{eq:small_l}).
$l(\theta^n;\mathcal{Q}_n)$ for each task $\mathcal{T}_{n}$ is obtained as below.
For each task $\mathcal{T}_{n}$ in $\mathcal{T}_{train}$, we use a support set $\mathcal{S}_n$ to learn a model by the learning algorihtm $F_{\phi}$.
The learned model is denoted as $\theta^n$, where $\theta^n=F_{\phi}(\mathcal{S}_n)$.
This procedure is equivalent to typical ML training. 
We called this step \textit{within-task training}.
Then $\theta^n$ is evaluated on $\mathcal{Q}_n$ to obtain $l(\theta^n;\mathcal{Q}_n)$ in (\ref{eq:meta-optimization}).
%This step is typical ML testing.
We called this step \textit{within-task testing}.
One execution of within-task training and followed by one execution of within-task testing is called an \textit{episode}. 

The optimization task below is solved to learn meta-parameteres $\phi$.
\begin{equation}
\phi^* = \arg \min_{\phi} L(\phi;\mathcal{T}_{train}). \label{eq:across-task_train}
\end{equation}
If $\phi$ is differentiable with respect to $L(\phi;\mathcal{T}_{train})$, then we can use gradient descent to learn meta-parameters; if not, we can use reinforcement learning algorithm or evolutionary algorithm. 
Solving (\ref{eq:across-task_train}) is called \textit{cross-task training} in this paper, which usually involves running many episodes on meta-training tasks.
To evaluate $\phi^*$, we need \textit{meta-testing tasks} $\mathcal{T}_{test}$, tasks for evaluating algorithms parameterized by meta-parameters $\phi^*$\footnote{If the learning processing of $\phi$ also involve some hyperperparameter selection, then \textit{meta-validation tasks} are needed, but in this paper, we ignore the discussion of meta-validation tasks for simplicity.}.
We do \textit{cross-task testing} on $\mathcal{T}_{test}$, that is, running an episode on each meta-testing task to evaluate algorithms parameterized by meta-parameters $\phi^*$.

In order to facilitate the reading of our paper, we summarize the most important terminologies and their meanings in Table~\ref{table:Terminologies} in the appendix.

\section{Task Construction} \label{sec:task}
In this section, we discuss different settings of constructing meta-training tasks $\mathcal{T}_{train}$ and meta-testing tasks $\mathcal{T}_{test}$.

\subsection{Cross-domain Transfer} \label{sec:cross-domain}
A typical setting for constructing the tasks is based on domains~\cite{Qian:ACL19,yan-etal-2020-multi-source,Li_Wang_Yu_2020,unsupervisedMT-acl21,chen-etal-2020-low,Huang:ACL20,dai-etal-2020-learning,wang-etal-2021-variance,dingliwal-etal-2021-shot,ST-dialogue-AAAI21}.
In this setting, all the tasks, no matter belonging to $\mathcal{T}_{train}$ or $\mathcal{T}_{test}$, are the same NLP problems.
In each task $\mathcal{T}_n$, the support set $\mathcal{S}_n$ and the query set $\mathcal{Q}_n$ are from the same domain, while different tasks contain the examples from different domains.
In each task, the model is trained on the support set of a domain (usually having a small size) and evaluated on the query set in the same domain, which can be considered as \textit{domain adaptation}.
From the meta-training tasks $\mathcal{T}_{train}$, cross-task training finds meta-parameters $\phi^*$ parameterizing the learning algorithm $F_{\phi^*}$.
With a sufficient number of tasks in $\mathcal{T}_{train}$, cross-task training should find a suitable $\phi^*$ for a wide range of domains, and thus also works well on the tasks in $\mathcal{T}_{test}$ containing the domains unseen during cross-task training.
Hence, meta-learning can be considered as one way to improve \textit{domain adaptation}.
If the support set in each task includes only a few examples, the meta-learning has to find the meta-parameters $\phi^*$ that can learn from a small support set and generalize well to the query set in the same domain.
Therefore, meta-learning is considered one way to achieve \textit{few-shot learning}.

The cross-domain setting is widespread.
We only provide a few examples in this subsection.
In MT, each meta-training task includes the documents from a specific domain (e.g., news, laws, etc.), while each meta-testing task also contains documents from one domain but not covered by the meta-training tasks (e.g., medical records)~\cite{Li_Wang_Yu_2020}. 
For another example, both meta-training and meta-testing tasks are DST.
The meta-training tasks include hotel booking, flight ticket booking, etc., while the testing task is taxi booking~\cite{Huang:ACL20,wang-etal-2021-variance,dingliwal-etal-2021-shot}. 
Domain has different meanings in different NLP problems. 
For example, in speech processing tasks, the domains can refer to accents~\cite{Winata:INTERSPEECH2020,SPmetaNLP21} or speakers~\cite{Klejch:ASRU19,SP-ICASSP21,huang2021metatts}. 

%There are more than 7,000 languages spoken in the world, over 90 of which have more than 10 million native speakers each (Eberhard et al., 2019). Despite this, very few languages have proper linguistic resources when it comes to natural language understanding tasks (Joshi et al., 2020). 

\subsection{Cross-lingual Transfer}   
\label{sec:cross-lingual}

If we consider different languages as different domains, then the cross-lingual transfer can be regarded as a special case of cross-domain transfer.
Suppose each task contains the examples of an NLP problem from one language, and different tasks are in different languages. 
In this case, cross-task training finds meta-parameters $\phi^*$ from the languages in $\mathcal{T}_{train}$, and cross-task testing evaluate the meta-parameters $\phi^*$ on new langauges in $\mathcal{T}_{test}$.
This setting aims at finding the learning algorithm $F_{\phi^*}(.)$ that works well on the NLP problem of any language given the support set of the language. 
Cross-language settings have been applied to NLI and QA in X-MAML \cite{nooralahzadeh-etal-2020-zero}, documentation classification~\cite{van-der-heijden-etal-2021-multilingual}, dependency parsing~\citep{multiparsing}, MT~\citep{Gu:EMNLP18}, and ASR~\citep{Hsu:ICASSP20,Winata:ACL2020,Chen:INTERSPEECH20,ASR-sample-AAAI21}.

For the meta-learning methods aiming at learning the initial parameters like MAML (will be introduced in Section~\ref{subsec:INIT}), the network architecture used in all tasks must have the same network architecture.
A unified network architecture across all tasks is not obvious in cross-lingual learning because the vocabularies in different tasks are different.
Before multilingual pretrained models are available, unified word embeddings across languages are required.
\citet{Gu:EMNLP18} uses the universal lexical representation to overcome the input-output mismatch across different languages.
Recently, by using multilingual pretrained models as encoders, such as M-BERT~\cite{devlin-etal-2019-bert} or XLM-R~\cite{conneau-etal-2020-unsupervised}, all languages can share the same network architecture~\cite{nooralahzadeh-etal-2020-zero,van-der-heijden-etal-2021-multilingual}.

\subsection{Cross-problem Training} % (Hung-yi) 
\label{sec:cross-problem}
Here the meta-training and meta-testing tasks can come from different problems.
For example, the meta-training tasks include MT and NLI, while the meta-testing tasks include QA and DST.
The cross-problem setting is not usual, but there are still some examples. 
In~\citet{Bansal:arXiv19}, the meta-training tasks are the GLUE benchmark tasks~\cite{wang-etal-2018-glue}, while the meta-testing tasks are NLP problems, including entity typing, NLI, sentiment classification, and various other text classification tasks, not in the GLUE. 
All the meta-training and meta-testing tasks can be formulated as classification but with different classes.
In~\citet{Indurthi:arXiv19}, the meta-training tasks are MT and ASR, while the meta-testing task is speech translation (ST). 
\textcolor{black}{CrossFit is a benchmark corpus for this cross-problem setting~\cite{ye-etal-2021-crossfit}.}
%Should I say more here?

The intrinsic challenge in the cross-problem setting is that different NLP problems may need very different meta-parameters in learning algorithms, so it may be challenging to find unified meta-parameters on the meta-training tasks that can generalize to meta-testing tasks.
In addition, the meta-learning algorithms learning initial parameters such as MAML require all the tasks to have a unified network architecture. 
If different problems need different network architecture, then the original MAML cannot be used in the cross-problem setting. 
LEOPARD~\citep{Bansal:arXiv19} and ProtoMAML~\citep{van-der-heijden-etal-2021-multilingual} are the MAML variants that can be used in the classification tasks with different class numbers.   
Both approaches use the data of a class to generate the class-specific head, so only the parameters of the head parameter generation model are required.
The head parameter generation model is shared across all classes, so the network architecture becomes class-number agnostic. 
On the other hand, recently, universal models for a wide range of NLP problems have been emgered~\citep{raffel2019exploring,chen2021speechnet,ao2021speecht5}. 
We believe the development of the universal models will intrigue the cross-problem setting in meta-learning.

\begin{figure}
    \centering
    \includegraphics[width=0.50\textwidth]{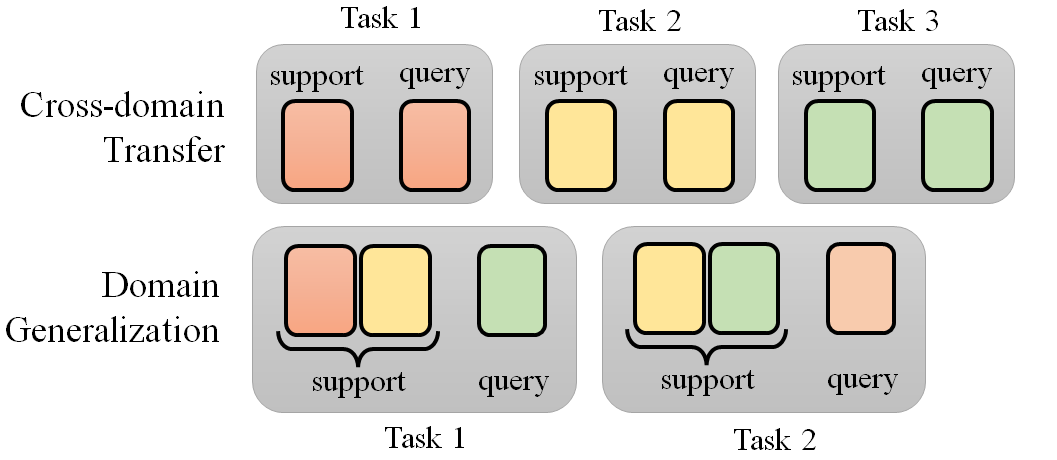}
    \caption{The task construction of cross-domain tranfer in Section~\ref{sec:cross-domain} and domain generalization in Section~\ref{subsec:DG}. Different colors represents data from different domains.  }
    \label{fig:DG}
\end{figure}

%\begin{itemize}
%\item In LEOPARD~\citep{Bansal:arXiv19}, the meta-training tasks are the GLUE benchmark tasks, while the meta-testing tasks are NLP problems, including entity typing, natural language inference, sentiment classification, and various other text classification tasks, not in the GLUE. 
%All the meta-training and meta-testing tasks can be formulated as classification but with different classes.
%\item The meta-training tasks are MT and automatic speech recognition (ASR), while the meta-testing task is speech translation (ST)~\citep{Indurthi:arXiv19}. Here the meta-training and meta-testing tasks have different input/output modalities. 
%\end{itemize}

\subsection{Domain Generalization} \label{subsec:DG}
Traditional supervised learning assumes that the training and testing data have the same distribution. 
Domain shift refers to the problem that a model performs poorly when training data and testing data have very different statistics. 
\textcolor{black}{
Domain adaptation in Section~\ref{sec:cross-domain} uses little domain-specific data to adapt the model\footnote{The domain-specific data are usually labelled, but unlabelled domain-specific data can be leveraged as well~\cite{UnsuperviedDomainSurvey}, which is out of scope here.}.}
On the other hand, \textit{domain generalization} techniques attempt to alleviate the domain mismatch issue by producing models that generalize well to novel testing domains. 

Meta-learning can also be used to realize domain generalization by learning an algorithm that can train from one domain but evaluate on the other. 
To simulate the domain generalization scenario, a set of meta-training tasks are constructed by sampling data from different domains as the support and query sets.
With the meta-training tasks above, cross-task training will find the meta-parameters $\phi^*$ that work well on the scenario where the training (support) and testing (query) examples are from different domains. 
Fig.~\ref{fig:DG} shows how to construct tasks for domain generalization and compares the construction with the cross-domain transfer setting.
The setting has been used to improve the domain generalization for semantic parsing~\cite{wang-etal-2021-meta} and language generalization\footnote{if a language is considered as a domain} for sentiment classification and relevance classification~\citep{li-etal-2020-learn}.

\subsection{Task Augmentation} 
\label{sec:task-generation}

In meta-learning, it is critical to have a large number of diverse tasks in the meta-training tasks $\mathcal{T}_{train}$ to find a set of meta-parameters $\phi^*$ that can generalize well to the meta-testing tasks.
However, considering the setting in the previous subsections, different tasks contain examples in various domains, language, or even NLP problems, so a large and diverse $\mathcal{T}_{train}$ are often not available.
In typical ML, data augmentation comes in handy when data is lacking.
In meta-learning, augmenting tasks is similarly understood as data augmentation in ML. 
Data augmentation becomes task augmentation because the ``training examples'' in meta-learning are a collection of tasks.
Task augmentation approaches in meta-learning can be categorized into two main directions: a) Inventing more tasks (without human labeling efforts) to increase the number and diversity of the meta-training tasks $\mathcal{T}_{train}$. b) Splitting training data from one single dataset into homogenous partitions that allow applying meta-learning techniques and therefore improve the performance.
NLP-specific methods have been proposed in both categories.

\paragraph{Inventing more tasks}
The main question is how to construct a massive amount of tasks efficiently.
There is already some general task augmentation approahces proposed for general meta-learning~\cite{pmlr-v139-yao21b,pmlr-v139-ni21a,NEURIPS2020_3e5190ee,yao2021metalearning}.
Here we only focus on NLP-specific approaches.
Inspired from the self-supervised learning,~\citet{bansal-etal-2020-self} generates a large number of cloze tasks, which can be considered as multi-class classification tasks but obtained without labeling effort, to augment the meta-training tasks.
\textcolor{black}{
\citet{bansal-etal-2021-diverse} further explores the influence of unsupervised task distribution and creates task distributions that are inductive to better meta-training efficacy.
The self-supervised generated tasks improve the performance on a wide range of different meta-testing tasks which are classification problems~\cite{bansal-etal-2020-self}, and it even performs comparably with supervised meta-learning methods on FewRel 2.0 benchmark~\cite{gao-etal-2019-fewrel} on 5-shot evaluation~\cite{bansal-etal-2021-diverse}.}

\paragraph{Generating tasks from a monolithic corpus}
Many tasks can be constructed with one monolithic corpus (\citet{Huang:NAACL18,Guo:ACL19,Wu:EMNLP19,jiang2019improved, Chien:INTERSPEECH19,li2020learning,maclaughlin2020evaluating,wang2020hat,pasunuru2020fenas,learningToLearn:metaNLP21,murty-etal-2021-dreca}).
First, the training set of the corpus is split into support partition, $\mathcal{D}_s$, and query partition, $\mathcal{D}_q$. 
Two subsets of examples are sampled from $\mathcal{D}_s$ and $\mathcal{D}_q$ as the support set, $\mathcal{S}$, and query set, $\mathcal{Q}$, respectively. 
In each episode, model parameters $\theta$ are updated with $\mathcal{S}$, and then the losses are computed with the updated model and $\mathcal{Q}$.
The meta-parameters $\phi$ are then updated based on the losses, as the meta-learning framework introduced in Section~\ref{sec:background}.  
The test set of the corpus is used to build $\mathcal{T}_{test}$ for evaluation.
As compared to constructing $\mathcal{T}_{train}$ from multiple relevant corpora, which are often not available, building $\mathcal{T}_{train}$ with one corpus makes meta-learning methodology more applicable. 
Besides, results obtained from one corpus are more comparable with existing NLP studies. 
However, only using a single data stream makes the resulting models less generalizable to various attributes such as domains and languages.  

How to sample the data points to form a task\footnote{If a corpus includes data from different domains, and we sample the data in the same domain to create a task, then the setting here becomes cross-domain in Section~\ref{sec:cross-domain}.} is the key in such category.
In NAS research in Section~\ref{subsec:NAS}, the support and query sets are usually randomly sampled. 
Learning to Compare in Section~\ref{subsec:COM} splits the data points of different classes in different tasks based on some predefined criteria.
There are some NLP-specific ways to construct the tasks.
In~\citet{Huang:NAACL18}, a relevance function is designed to sample the support set $\mathcal{S}$ based on its relevance to the query set $\mathcal{Q}$.
In~\citet{Guo:ACL19}, a retrieval model is used to retrieve the support set $\mathcal{S}$ from the whole dataset.
DReCa~\cite{murty-etal-2021-dreca} applies clustering on BERT representations to create tasks.

%is a task augmentation strategy that takes as input a entire dataset, and then decomposes it to approximately recover some of the latent reasoning categories underlying the dataset, such as various syntactic constructs within a dataset, or semantic categories such as quantifiers and negation.

\section{Meta-Learning for NLP Tasks} \label{sec:specific}
This section shows the most popular meta-learning methods for NLP and how they fit into NLP tasks.
Due to space limitations, only the major trends are mentioned. 
Please refer to Table~\ref{table:methods} and~\ref{table:methods2} in the appendix for a complete survey. 

\subsection{Learning to Initialize}  \label{subsec:INIT}
In typical DL, gradient descent is widely used to solve (\ref{eq:optmization}).
Gradient descent starts from a set of initial parameters $\theta^0$, and then the parameters $\theta$ are updated iteratively according to the directions of the gradient.
There is a series of meta-learning approaches targeting at learning the initial parameters $\theta^0$.
In these learn-to-init approaches, the meta-parameters $\phi$ to be learned are the initial parameters $\theta^0$ for gradient descent, or $\phi=\theta^0$. 
MAML~\cite{Finn:ICML17} and its first-order approximation, FOMAML~\cite{Finn:ICML17}, Reptile~\cite{nichol2018firstorder}, etc., are the representative approaches of learn-to-init. 
We surveyed a large number of papers using MAML-based approaches to NLP applications in the last three years and summarized them in Table~\ref{table:maml} in the appendix.

\paragraph{Learning to Initialize v.s. Self-supervised Learning}
The learn-to-init approaches aim at learning a set of good initial parameters.
On the other hand, self-supervised approaches like BERT also have the same target.
There is a natural question: are they complementary? 
Based on the survey in Table~\ref{table:maml} in the appendix, it is common to use the self-supervised models to ``initialize'' the meta-parameters $\phi$ in learn-to-init approaches.
To find the optimal $\phi^*$ in (\ref{eq:meta-optimization}), gradient descent is used as well, and thus the ``initial parameters for initial parameters'', or $\phi^0$ is required. 
A self-supervised model usually serves the role of $\phi^0$, and the learn-to-init approaches further update $\phi^0$ to find $\phi^*$.

Learn-to-init and self-supervised learning are complementary. 
The self-supervised objectives are different from the objective of the target NLP problem, so there is a ``learning gap''. 
On the other hand, learn-to-init approaches learn to achieve good performance on the query sets of the meta-training tasks, so it directly optimizes the objective of the NLP problems. 
The benefit of self-supervised learning is that it does not require labeled data, while labeling is still needed to prepare the examples in meta-training tasks. 

\paragraph{Learning to Initialize v.s. Multi-task Learning}
Multi-task learning is another way to initialize model parameters, which usually serves as the baseline of learn-to-init in the literature.  
In multi-task learning, all the labelled data from the meta-training tasks is put together to train a model.
That is, all the support sets $\mathcal{S}_n$ and query sets $\mathcal{Q}_n$ in the meta-training tasks $\mathcal{T}_{train}$ are put together as a training set $\mathcal{D}$, and the loss (\ref{eq:optmization}) is optimized to find a parameter $\theta^*$.
Then $\theta^*$ is used as initial parameters for the meta-testing tasks. 

Both multi-task learning and meta-learning leverage the examples in the meta-training tasks, but with different training criteria. 
Learn-to-init finds the initial parameters suitable to be updated by updating the model on the support sets and then evaluating it on the query sets. 
In contrast, multi-task learning does not consider that the initial parameters would be further updated at all during training.
Therefore, in terms of performance, learn-to-init is usually shown to be better than multi-task learning~\cite{Dou:EMNLP19,chen-etal-2020-low}.
On the other hand, in terms of training speed, meta-learning, which optimizes (\ref{eq:meta-optimization}), is more computationally intensive than multi-task learning optimizing (\ref{eq:optmization}).

\paragraph{Three-stage Initialization}
Since learn-to-init, multi-task, self-supervised learning all have their pros and cons, they can be integrated to draw on the strong points of each other. 
A common way to integrate the three approaches is ``three-stage initialization'' as below.
a) First, initialize a model by self-supervised learning, which leverages unlabeled data. Its objective is usually not directly related to the target NLP problem. 
b) Then, multi-task learning is used to fine-tune the self-supervised model.
The objective of multi-task learning is the target NLP problem but does not consider the update procedure in gradient descent. 
c) Finally, learn-to-init, which finds the initial parameters suitable for update, is used to fine-tune the multi-task model. 

Learn-to-init is chosen to be the last stage because its training objective is closest to the target of looking for good initial parameters, but it is the most computationally intensive method, and thus it is only used to change the model a little bit.
The three-stage initialization has been tested in several works~\cite{nooralahzadeh-etal-2020-zero,SP-ICASSP21,van-der-heijden-etal-2021-multilingual,multiparsing}, but it does not always improve the performance~\cite{SP-ICASSP21,van-der-heijden-etal-2021-multilingual}. 

\textcolor{black}{
\paragraph{Challenges}
Learn-to-init is an essential paradigm for few-shot learning and usually achieves outstanding results in the few-shot learning benchmarks of image classification~\cite{Triantafillou2020Meta-Dataset}.
However, it has fallen short of yielding state-of-the-art results on NLP few-shot learning benchmarks~\cite{ye-etal-2021-crossfit,ICTACL2022,FLEX}. 
For example, on the cross-task few-shot learning benchmark, CrossFit, simple multi-task learning outperforms existing learn-to-init in many cases~\cite{ye-etal-2021-crossfit}.
One possible reason is meta-learning methods are susceptible to hyper-parameters and even random seeds~\cite{antoniou2018how}. 
Hence, it is difficult to obtain decent performance without exhaustively tuning hyperparameters.
The research about developing more stable learn-to-init methods may lead to more practical real-world applications for the approaches. 
There is a study about stabilizing the cross-task training of learn-to-init methods by reducing the variance of gradients for NLP~\cite{wang-etal-2021-variance}.
}

\subsection{Learning to Compare} \label{subsec:COM}
Learning to Compare methods are widely applied to NLP tasks. Among many others, we find applications of Learning to Compare methods in text classification \cite{Yu:ACL18,Tan:EMNLP19,Geng:EMNLP19, Sun:EMNLP19,geng2020dynamic}, sequence labeling \cite{Hou:ACL20,oguz2021few}, semantic relation classification \cite{Ye:ACL19, Chen:EMNLP19, Gao:AAAI19, ren2020two}, knowledege completion \cite{xiong2018one, wang2019tackling, zhang2020few, sheng2020adaptive} and speech recognition \cite{lux2021meta} tasks. 

Most of the proposed methods are based on Matching Network \cite{Vinyals:NIPS16}, Prototypical Network \cite{Snell:NIPS17} and Relation Network \cite{Sung:CVPR18}, and extend these architectures in two aspects: 
a) how to embed text input in a vector space with/without context information, and b) how to compute the distance/similarity/relation between two inputs in this space. Since these questions have had deep roots in the computation linguistics research for many years \cite{schutze1992dimensions, manning1999foundations}, Learning to Compare methods is one of the most important methods among other meta-learning methods in the context of NLP despite their simplicity. \textcolor{black}{
Notably, to date, such family of methods is mainly applied to classification tasks.}

\subsection{Neural Network Architecture Search} \label{subsec:NAS}
Neural network architecture search (NAS) is another common meta-learning technique applied to NLP including language modeling (WikiText-103 \citep{merity2017revisiting}, PTB \citep{mikolov2010recurrent}), NER (CoNLL-2003 \citep{sang2003introduction}), TC (GLUE \citep{wang2019glue}), and MT (WMT'14 \citep{bojar-EtAl:2014:W14-33}). As discussed in Section~\ref{sec:task-generation}, these techniques are often trained/evaluated with a single, matched dataset, which is different from other meta-learning approaches.

Moreover, in contrast to conventional NAS methods that focus on learning the topology in an individual recurrent or convolutional cell, NAS methods have to be redesigned in order to make the search space suitable for NLP problems, where contextual information often plays an important role. 
%NAS techniques in NLP are usually benchmarked on tasks including language modeling (WikiText-103 \citep{merity2017revisiting}, PTB \citep{mikolov2010recurrent}), NER (CoNLL-2003 \citep{sang2003introduction}), TC (GLUE \citep{wang2019glue}), and MT (WMT'14 \citep{bojar-EtAl:2014:W14-33}). As discussed in Section~\ref{sec:task-generation}, these techniques are trained/evaluated with a single, matched dataset, which is different from other meta-learning approaches. As the progress of meta-learning research, we expect more NLP problems and the condition of mismatched training/inference will be explored in the context of NAS.
\citet{jiang2019improved} pioneers the application of NAS to NLP tasks beyond language modeling (NER in this case), and improves differentiable NAS by redesigning its search space for natural language processing. \citet{li2020learning} extends the search space of NAS to cover more RNN architectures and allow the exploring of intra- and inter-token connection to increase the expressibility of searched networks. As the popularity of pre-trained language models (PLM) grows in NLP area, researchers also apply NAS to discover better topology for PLM such as BERT. \citet{wang2020hat} introduces Hardware-Aware Transformers (HAT) to search Transformer architecture optimized for inference speed and memory footprint in different hardware platforms. NAS-BERT \citep{xu2021bert} and AdaBERT \citep{ijcai2020-341} explores task-agnostic and task-dependent network compression techniques with NAS respectively. EfficientBERT \citep{dong-etal-2021-efficientbert-progressively} applies NAS to search for more efficient architecture of feed-forward network that is suitable for edge device deployment.

To show the efficacy of NAS, we summarize the performance of several state-of-the-art NAS approaches on GLUE benchmarks \citep{wang2019glue} in Table \ref{table:NAS_Performance_on_GLUE} in the appendix. These approaches are applied to BERT to discover architectures with smaller sizes, faster inference speed, and better model accuracy. For comparison, performance from original and manually compressed BERT models is also presented. The results show that the BERT architecture improved by NAS yields performance competitive to BERT (c.f., 82.3 from EfficientBERT vs 82.5 from BERT) and is 6.9x smaller and 4.4x faster. The searched architecture also outperforms manually designed, parameter- and inference-efficient model (MobileBERT\textsubscript{TINY}) at similar size and speed. These results suggest the efficacy of NAS in discovering more efficient network architectures. As NLP researchers continue to design even larger PLMs while the need of deployment on edge devices grows, we expect there will be increasing investment in innovating NAS techniques to make PLM networks more compact and accelerate inference.

\paragraph{Challenges}
The main bottleneck for NAS being widely applied is the prohibitive requirement in computation resources for architecture search. Approaches such as Efficient Neural Architecture Search (ENAS, \citet{pham2018efficient}) and Flexible and Expressive Neural Architecture Search (FENAS, \citet{pasunuru2020fenas}) are proposed to improve the search efficiency. 
As PLMs usually have bulky sizes and slow training speed, search efficiency is even more critical when applying NAS to PLM. 
Weight-sharing techniques are often applied to accelerate searching \cite{wang2020hat, dong-etal-2021-efficientbert-progressively, xu2021bert}.

%\citet{jiang2019improved} pioneers the application of NAS to NLP tasks beyond language modeling (NER in this case), and improves differentiable NAS by redesigning its search space for natural language processing. \citet{li2020learning} extends the search space of NAS to cover more RNN architectures and allow the exploring of intra- and inter-token connection to increase the expressibility of searched networks. \citet{pasunuru2020fenas} proposes Flexible and Expressive Neural Architecture Search (FENAS) to expand the search space of Efficient Neural Architecture Search (ENAS, \citet{pham2018efficient}) with more functions (e.g., skip-based tanh, ReLU) and atomic-level operations (e.g., addition, subtraction, element-wise multiplication). As the popularity of Transformer architecture grows in NLP area, researchers also apply NAS to discover better topology for Transformer. \citet{wang2020hat} introduces Hardware-Aware Transformers (HAT) to search Transformer architecture optimized for inference speed and memory footprint in different hardware platforms. As Transformer is heavy in computation, weight-sharing technique is adopted to make architecture search more efficient.

\subsection{Meta-learning for Data Selection}

Multi-linguality, multi-task, and multi-label see many impacts on NLP problems due to the diversity of human languages. 
To learn models with balanced performance over attributes (e.g., languages, tasks, labels), a common approach is to weight the training examples for data selection to learn models with balanced performance over the attributes, and it is a natural assumption that meta-learning techniques derive more generalizable weighting than manually tuned hyperparameters. 
For example, \citet{Wu:EMNLP19} add another gradient update step wrapping the conventional classifier update for training meta-parameters that controls the weight when aggregating losses from different labels to update classifier's parameters.
%\citet{learningToOptimize:metaNLP21} proposes a meta-optimizer to soft-select portion of parameters in pre-trained networks to be frozen during fine-tuning. 
In addition to gradient update, meta-learned weights are also applied directly to training examples for data selection to address the issue of noisy labeling. \citet{shu2019meta} propose a technique to jointly learn a classifier and a weighting function, where a conventional gradient update for the classifier and a meta-learning update for the weighting is performed alternatively. The function weights examples to mitigate model overfitting towards biased training data caused by corrupted labels or class imbalance. \citet{label-correction-aaai21} apply a similar framework but extend the weighting with a label correction model. Both techniques show improvement over SOTA in text classification with biased training data. 

Additionally, as the progress in the research of pre-training and transfer learning, there is a trend of leveraging datasets in multiple languages, domains, or tasks to jointly pre-train models to learn transferable knowledge. A meta-learned data selector can also help in this scenario by choosing examples that benefit model training and transferability. For instance, \citet{wang2020balancing} investigate the common challenge of imbalanced training examples across languages in multilingual MT, which is conventionally addressed by tuning hyperparameters manually to up-sample languages with less resources. The authors propose Differentiable Data Selection (DDS) to parameterize the sampling strategies. DDS is trained with episodes and REINFORCE algorithm to optimize parameters of sampler and MT models in an alternating way for the MT models to converge with better performance across languages. 
\citet{pham2020meta} formulate data sampling for multilingual MT as a problem of back-translation to generate examples of parallel utterances from unlabeled corpora in target language. The back-translation is jointly trained with MT models to improve translation result through better distribution of training examples and data augmentation. \citet{tarunesh2021meta} further study knowledge transferring across tasks and languages. The authors combine Reptile and DDS to meta-learn samplers with six different languages (en, hi, es, de, fr, and zh) and five different tasks (QA, NLI, paraphrase identification, POS tagging, and NER) and demonstrate competitive performance on XTREME multilingual benchmark dataset \citep{hu2020xtreme}.

%and more details will be discussed in Section \ref{subsec: learn_to_select}. 
%Table \ref{table:dataset-multi-lingual-task-label} summarizes common datasets for benchmarking meta-learning techniques described here.

%\begin{table}[tb!]
%\centering
%  \small
%  \caption{Datasets to benchmark meta-learning techniques under multi-linguality, multi-task, or multi-label conditions. We use the following abbreviations for tasks. \textbf{NLI}: Natural language inference. \textbf{MT}: Machine translation. \textbf{QA}: Question answering. \textbf{PI}: Paraphrase identification. \textbf{POS}: Part of speech tagging. \textbf{NER}: Named entity recognition. \textbf{EC}: Entity type classification. \textbf{TC}: Text classification.}
%  \label{table:dataset-multi-lingual-task-label}
%  \begin{tabular}{l|ll}
%    \toprule
%    Datasets & Attributes & Tasks \\ \midrule
%    XNLI \citep{conneau2018xnli} & multi-lingual & NLI \\  
%    \citep{qi2018and} & multi-lingual & MT \\  
%    XTREME & multi-lingual & QA, NLI, PI \\ 
%    $\;\;\;$ \citep{hu2020xtreme} & $\;\;\;$ \& task & $\;\;\;$ POS, NER \\ 
%    \makecell{FIGER/OntoNotes/BBN \\       $\;\;\;$ \citep{Wu:EMNLP19}} & \multirow{1}{*}[2mm]{multi-label} & \multirow{1}{*}[2mm]{EC} \\  
%    \makecell{RCV1-V2/Reuters-21578 \\ $\;\;\;$ \citep{Wu:EMNLP19}} & \multirow{1}{*}[2mm]{multi-label} & \multirow{1}{*}[2mm]{TC} \\
%    \bottomrule
%  \end{tabular}
%\end{table}

%\section{Benchmark} \label{sec_benchmark}
%Some numbers from popular datasets such as GLUE semEval etc.

\section{Meta-learning beyond Accuracy} \label{sec:advance}
In the previous sections, meta-learning is used to obtain better evaluation metrics for NLP applications. %usually in the context of domain adaptation.
This section illustrates how meta-learning can improve NLP applications from more aspects beyond performance.

\subsection{Learn to Knowledge Distillation}
Knowledge distillation method was proposed in \cite{hinton2015distilling}. 
The main goal is to transfer knowledge from a so-called teacher model, e.g., a vast neural network trained with a lot of training data, to a more compact student model, e.g., a neural network with much less trainable parameters.
The main weaknesses of this method are as follows: a) the number of teacher models is fixed to one that could limit the power of the transferring process; b) the teacher model is not optimized for the transferring process and c) the teacher model is not aware of the student model during the transferring process.
Meta-learning methods can be applied to partially fix these issues. The high-level idea is to increase the number of teacher models and the number of student models and consider each pair of a teacher model and a student model as a task in the meta-learning framework. By doing so, we can train a meta teacher model that works better than a single teacher model~\cite{pan2020meta}, and we can optimize the transferring process and force the teacher model to be aware of the student model \cite{zhou2021meta}.  % zhou2021meta is accepted by ACL

\subsection{Learn to Life-long learning} \label{subsec: learn_to_lifelong} %(Hung-yi)
This subsection discusses how to use meta-learning to improve \textit{lifelong learning} (LLL) \cite{Chen18-llml}.
The real world is changing and evolving from time to time, and therefore machines naturally need to update and adapt to the new data they receive.
However, when a trained deep neural network is adapted to a new dataset with a different distribution, it often loses the knowledge previously acquired and performs the previous seen data worse than before.
This phenomenon is called \textit{catastrophic forgetting}~\cite{Mccloskey89-catastrophic}.
%LLL aiming for training a single model to perform a stream of tasks without forgetting those learned earlier becomes a necessary goal for the continuously changing real world.
There is a wide range of LLL approaches aiming for solving catastrophic forgetting~\cite{LLLReview19}.  
Among them, the following directions apply meta-learning:
%This subsection focuses only on the LLL approaches that can be further improved by meta-learning
\footnote{On the other hand, in meta-learning, usually, we assume stationary task distribution. 
Can we do meta-learning with evident distributional shift or when tasks arrive sequentially?
There is also research along the direction~\cite{finn20191onlinemeta,yap2021onlinemeta}, but out of the scope of this review paper.}

\paragraph{Meta-learning for Regularization-based LLL methods}
Regularization-based LLL methods aim to consolidate essential parameters in a model when adapting models with new data~\cite{Kirkpatrick17-ewc, Zenke17-si, Schwarz18-online-ewc,  Aljundi18-mas, Ehret20-rnn-hypernet}. % by adding regularization terms in the loss function
Meta-learning targets ``how to consolidate'' and has some successful examples in NLP applications. 
KnowledgeEditor~\cite{decao2021editing} learns the parameter update strategies that can learn the new data and simultaneously retain the same predictions on the old data.
KnowledgeEditor has been applied to fact-checking and QA.
Editable Training~\cite{sinitsin2020editing} employs learn-to-init approaches to find the set of initial parameters, ensuring that new knowledge can be learned after updates without harming the performance of old data.
Editable Training empirically demonstrates the effectiveness on MT.

\paragraph{Meta-learning for Data-based LLL Methods}
The basic idea of data-based methods is to store a limited number of previously seen training examples in memory and then use them for empirical replay, that is, training on seen examples to recover knowledge learned~\cite{sprechmann2018memorybased,EpisodicMemoryLLLL,Sun19-lamol} or to derive optimization constraints~\cite{Lopez17-gem,Li17-lwf,Saha21-gpm}.
A hurdle for data-based approaches is the need to store an unrealistically large number of training examples in memory to achieve good performance.
To achieve sample efficiency, \citet{obamuyide-vlachos-2019-meta,wang-etal-2020-efficient,wu2021meta-lifelong-re} uses meta-learning to learn a better adaptation algorithm that recovers the knowledge learned with a limited amount of previously seen data. 
Experiments on text classification and QA benchmarks validate the effectiveness of the framework, achieving state-of-the-art performance using only 1\% of the memory size~\cite{wang-etal-2020-efficient}.

\section{Conclusion}
This paper investigates how meta-learning is used in NLP applications.
We review the task construction settings (Section~\ref{sec:task}), the commonly used methods including learning to initialize, learning to compare and neural architecture search (Section~\ref{sec:specific}), and highlight research directions that go beyond improving performance (Section~\ref{sec:advance}).
We hope this paper will encourage more researchers in the NLP community to work on meta-learning.

%We found that the NLP community lacks a standard benchmark for meta-learning.
%In CV, meta-learning algorithms can be evaluated and compared on benchmarks such as Omniglot~\cite{Omniglot}, miniImageNet~\cite{Vinyals:NIPS16}, etc.
%However, in NLP, there is currently no such widely accepted benchmark, which hinders development.
%We believe that an easy access benchmark for evaluating meta-learning in NLP applications is a critical next step to further advances in meta-learning.

% DANIEL: move the rest to appendix for clarity. Some paragraphs are pretty useful and can be considered in final papers.

%Hung-yi: All the tables are in the appendix.

\footnotesize
\bibliography{eacl2021}
\bibliographystyle{acl_natbib}

\clearpage
\newpage
\onecolumn
\appendix

\input{appendix}

\end{document}

%% file: appendix.tex
\section{Appendx}

\begin{table*}[ht]
\caption{Terminologies and their meanings.}
\label{table:Terminologies}
\begin{center}
\small
\begin{tabular}{ |l|l| } 
 \hline 
Terminologies & Meaning\\
 \hline \hline
 (NLP) Problem & a type of NLP problems like QA, POS, or MT \\
 Model Parameter & parameters of models making inference for underlying problems \\
 Meta-parameter & parameters of learning algorithms (e.g., model init, optimizers) that are shared across tasks \\
 Support Set & a set of training examples for updating model parameters  \\
 Query Set & a set of testing examples for evaluating model parameters \\
 Task & combination of one support set and one query set\\
 Within-task Training & learning model parameter with support set \\
 Within-task Testing & using query set to evaluate model parameters \\ 
 Episode & one execution of within-task training and followed by one execution of within-task testing \\
 Meta-training Tasks & tasks generated for learning meta-parameter \\
 Meta-testing Tasks & tasks generated for evaluating algorithms parameterized by meta-parameter \\
 Cross-task Training & learning meta-parameter, which usually involves running many episodes on meta-training tasks \\
 Cross-task Testing &  running an episode on each meta-testing task to evaluate algorithms parameterized by meta-parameter \\
\hline 
\end{tabular}
\end{center}
\end{table*}

\begin{table*}[h!]
  \centering
  \small
  \caption{An organization of works on meta-learning in NLP. 
  The \textbf{Application} column lists the applications that are performed in corresponding papers. We use the following abbreviations.  \textbf{QA}: Question Answering.  \textbf{MT}: Machine Translation. \textbf{TC}: Text Classification (including Natural Langauge Inference). \textbf{IE}: Information Extraction (including Relation Classificaiton and Knowledge Graph Completion). \textbf{WE}: Word Ebedding  \textbf{TAG}: Sequence Tagging. \textbf{PAR}: Parsing. \textbf{DST}: Dialgoue State Tracking. \textbf{DG}: Dialgoue Generation (including  Natural Language Generation). \textbf{MG}: Multimodal Grounding. \textbf{ASR}: Automatic Speech Recognition. \textbf{SS}: Source Separation. \textbf{KS}: Keyword Spotting. \textbf{VC}: Voice Cloning. \textbf{SED}: Sound Event Detection.  
 The \textbf{Method} column lists the involving meta-learning methods. \textbf{INIT} is learning to initialize; \textbf{COM} is learning to compare; \textbf{NAS} is network architecture search; \textbf{OPT} is learning to optimize; \textbf{ALG} is learning the learning algorithm; \textbf{SEL} is learning to select data. 
  \textbf{Task construction} column lists the way each work is built for training meta-parameters. Please refer to Section~\ref{sec:task} for the description about task construction. 
  }
  \label{table:methods}
  \resizebox{\textwidth}{92mm}{
  \begin{tabular}{llll}
    \toprule
    Work & Method & Application  &  Task construction  \\ 
    \midrule
 \cite{Dou:EMNLP19} & INIT & TC & Cross-problem \\ %吳元魁 Y
 \cite{Bansal:arXiv19}   & INIT & TC & Cross-problem  \\  %吳元魁 Y
 \cite{holla-etal-2020-learning}& INIT & TC & A task includes sentences containing the same word with different senses. \\ %Cross-domain\footnote[1]{Each word is considered as a ``domain''.} \\ %吳元魁 Y
 \cite{zhou2021meta}& INIT & TC & Knowledge Distallation \\ %吳元魁 Y
 \cite{pan2020meta}& COM      & TC & Knowledge Distallation \\ %謝濬丞 Y
  \cite{van-der-heijden-etal-2021-multilingual}& INIT & TC & Cross-lingual  \\ %謝濬丞 Y
 \cite{bansal-etal-2020-self}& INIT & TC & Cross-problem (some tasks are generated in an self-supervised way) \\%謝濬丞 Y
 \cite{murty-etal-2021-dreca}& INIT & TC & Cross-problem \\ %using data augmentation %謝濬丞 Y
  \cite{wang-etal-2021-variance} & INIT & TC, DST & Cross-domain \\%謝濬丞 Y
  \cite{Yu:ACL18}&   COM &TC & Cross-domain \\%張凱為 Y
 \cite{Tan:EMNLP19}&  COM &TC & Cross-domain \\%張凱為 Y
 \cite{Geng:EMNLP19}&   COM &TC & Cross-domain\\%張凱為 Y
 \cite{Sun:EMNLP19}&  COM &TC & The tasks are seperated by class labels. \\%張凱為 Y
 \cite{geng2020dynamic}&   COM &TC & The tasks are seperated by class labels. \\%張凱為
  \cite{li-etal-2020-learn} & COM &TC & Domain Generalization\\%蔡翔陞 Y
  \cite{Wu:EMNLP19}&  OPT & TC & Monolithic \\%蔡翔陞 Y
 \cite{pasunuru2020fenas} & NAS &TC & Monolithic \\%蔡翔陞 Y
 \cite{pasunuru2019continual} & NAS &TC & Monolithic \\%蔡翔陞 Y
 \cite{learningToOptimize:metaNLP21} & OPT &TC &Domain Generalization \\%蔡翔陞 Y
 \cite{label-correction-aaai21} & SEL  &TC & Monolithic \\%張致強Y
  \cite{Wu:AAAI20} & INIT & TAG & Cross-lingual \\ %Monolithic \\%張致強Y
 \cite{xia-etal-2021-metaxl} & INIT & TC, TAG & Cross-lingual \\ %Domain Generalization \\%張致強Y
 \cite{Hou:ACL20}  & COM & TAG & Cross-domain \\%張致強Y
 
 %\cite{yang2020frustratingly} & COM & TAG & Monolithic\footnotemark[3] \\%張致強Y
 
 \cite{oguz2021few} & COM & TAG & The tasks are seperated by class labels. \\%黃冠博Y
 \cite{li2020learning} & NAS  & DG & Monolithic  \\%黃冠博Y
 \cite{jiang2019improved} & NAS & TAG & Monolithic \\%黃冠博Y
  \cite{Obamuyide:ACL19} & INIT & IE & Each task includes the examples for a relation.  \\ % relation%黃冠博 classification Y
 \cite{Bose:arXiv19} & INIT & IE & Each task is a graph. \\ %黃冠博 
 \cite{Lv:EMNLP19} & INIT & IE & Each task includes the examples for a relation.  \\  %曾韋誠 Y
 \cite{Chen:EMNLP19}& COM & IE & Each task includes the examples for a relation.  \\%曾韋誠 Y
 %\cite{Xiong:EMNLP18}& COM & IE & Cross-domain\footnotemark[4]  \\%曾韋誠 Y
 \cite{Gao:AAAI19}& COM & IE & Each task includes the examples for a relation. \\%曾韋誠 Y
  \cite{ren2020two} & COM & IE &Each task includes the examples for a relation. \\%曾韋誠 Y
  \cite{xiong2018one}& COM & IE &Each task includes the examples for a relation. \\%陳建成 Y
 \cite{wang2019tackling}& INIT & IE &Each task includes the examples for a relation. \\%陳建成 Y
 \cite{zhang2020few}& COM & IE &Each task includes the examples for a relation.  \\%陳建成 Y
 \cite{sheng2020adaptive}& COM & IE &Each task includes the examples for a relation. \\%陳建成 Y
  \cite{Hu:ACL19} & INIT & WE & Each task includes the context of a word. \\%陳建成 Y
 \cite{Sun:EMNLP18} & COM & WE & Each task includes the context of a word. \\%高瑋聰 Y
 %\cite{li2020learning} & NAS & WE & \\%高瑋聰 Y
 % \cite{jiang2019improved}& NAS & WE & \\%高瑋聰 Y
  \cite{mhamdi-etal-2021-x} & INIT & QA, TAG & Cross-lingual, Domain Generalization\\%高瑋聰 Y
 \cite{nooralahzadeh-etal-2020-zero} &INIT & QA, TC & Cross-lingual \\%高瑋聰 Y
 \cite{yan-etal-2020-multi-source} & INIT & QA & Cross-domain \\%黃世丞 Y
  \cite{Gu:EMNLP18} & INIT & MT & Cross-lingual \\%黃世丞 Y
 \cite{Indurthi:arXiv19} & INIT & MT & Cross-problem \\%黃世丞 Y
 \cite{Li_Wang_Yu_2020} & INIT & MT & Cross-domain  \\%黃世丞 Y
 \cite{unsupervisedMT-acl21}  & INIT & MT & Cross-domain  \\%黃世丞 Y
  \cite{wang2020balancing} & SEL & MT & Monolithic\\%許湛然Y

     \hline
    \bottomrule
  \end{tabular}}
\end{table*}

\begin{table*}[h!]
  \centering
  \small
  \caption{Continue of Table~\ref{table:methods}. \citet{pham2020meta} learns a backtranslation model for data augmentation, so it is considered as SEL.}
  \label{table:methods2}
\resizebox{\textwidth}{88mm}{
  \begin{tabular}{llll}
    \toprule
    Work & Method & Application  &  Task construction  \\ 
    \midrule
     \cite{pham2020meta} & SEL & MT & Monolithic  \\%許湛然 %\footnote{Learn to do data augmentation} 
     \cite{Guo:ACL19}&INIT&PAR & Monolithic\\ %許湛然
\cite{Huang:NAACL18}&INIT&PAR & Monolithic\\%許湛然
\cite{multiparsing}&INIT&PAR & Cross-lingual \\%許湛然
\cite{chen-etal-2020-low}&INIT&PAR & Cross-domain \\%門玉仁Y
\cite{wang-etal-2021-meta}&INIT&PAR & Domain Generalization\\%門玉仁 Y
\cite{Qian:ACL19}&INIT & DG & Cross-domain  \\%門玉仁 Y
\cite{Madotto:ACL19}&INIT & DG & Cross-domain (each domain is one type of persona)  \\%門玉仁 Y
\cite{Mi:IJCAI19}&INIT& DG & Cross-domain\\%門玉仁 Y
\cite{Huang:ACL20}&INIT& DST &Cross-domain \\%孟妍 Y
\cite{dingliwal-etal-2021-shot}&INIT& DST &Cross-domain \\%孟妍 Y
\cite{huang-etal-2020-towards-low}&INIT& DST & Cross-domain\\%孟妍 Y
\cite{dai-etal-2020-learning}&INIT& DG &Cross-domain \\%孟妍 Y
\cite{ST-dialogue-AAAI21}&INIT& DG &Cross-domain \\%孟妍 Y
\cite{Chien:INTERSPEECH19}&  OPT & DG & Monolithic \\%馮子軒Y
 \cite{Hsu:ICASSP20}&INIT&ASR& Cross-lingual \\%馮子軒Y
\cite{Klejch:ASRU19}&INIT&ASR& Cross-domain (each domain refers to a speaker) \\%馮子軒Y
\cite{Winata:ACL2020}&INIT&ASR&Cross-lingual \\%馮子軒Y
\cite{Winata:INTERSPEECH2020}&INIT&ASR& Cross-domain (each domain refers to a accent) \\%馮子軒Y
\cite{ASR-sample-AAAI21}&INIT&ASR&  Cross-lingual\\%陳宣叡 Y 方法也許可以改成MAML -> MAML/FOMAML/Reptile，因為在ablation那邊有把它提出的方法(AMS)用在這幾種不同的方法上面
\cite{Klejch:INTERSPEECH18} & OPT&ASR& Cross-domain (each domain refers to a speaker)\\%陳宣叡 Y
\cite{Chen:INTERSPEECH20} & NAS&ASR&  Cross-lingual\\%陳宣叡 Y
\cite{Baruwa:IJSER19} & NAS &ASR& Monolithic \\%陳宣叡 Y 這邊method應該是打錯了 NSA -> NAS %感謝!
\cite{SP-ICASSP21} & INIT & SS & Cross-domain (each domain refers to a speaker)\\%陳宣叡 Y
\cite{SPmetaNLP21} & INIT & SS &  Cross-domain(each domain refers to a accent)\\%江宥呈 Y
\cite{Chen:arXiv18} &    INIT & KS & The tasks are separated by keyword sets.  \\%江宥呈 Y
\cite{parnami2020fewshot} & COM & KS & The tasks are separated by keyword sets.  \\%江宥呈 Y
\cite{KSSLT21}  & COM & KS &  The tasks are separated by keyword sets.  \\%江宥呈 Y
\cite{Mazzawi:INTERSPEECH19} & NAS & KS & Monolithic \\%江宥呈 Y
\cite{lux2021meta} & COM & KS &  The tasks are separated by keyword sets.  \\%張恆瑞 Y
 \cite{Serra:NeurIPS19} & ALG & VC & Cross-domain (each domain refers to a speaker)\\%張恆瑞 Y
 \cite{Chen:ICLR19}& ALG & VC & Cross-domain (each domain refers to a speaker)\\%張恆瑞 Y
 \cite{huang2021metatts} & INIT & VC & Cross-domain (each domain refers to a speaker)\\%張恆瑞 Y
\cite{tarunesh2021meta} & INIT, SEL & QA, TC, TAG & Cross-lingual, Cross-problem \\ % Our chosen tasks are question answering (QA), natural language inference (NLI), paraphrase identification (PA), part-of-speech tagging (POS) and named entity recognition (NER). %張恆瑞 Y
\cite{Eloff:ICASSP19} & COM & MG & Monolithic  \\ %黃維坪 Y
\cite{Suris:arXiv19} & ALG & MG & Each task contains multiple examples of text-image pairs. \\ %黃維坪 Y
\cite{learningToLearn:metaNLP21}& COM & MG &  Each task contains an image and a word set. \\ %黃維坪 Y

\cite{decao2021editing}   & OPT & TC, QA &  Life-long learning \\
\cite{sinitsin2020editing}& INIT  & MT &  Life-long learning \\

\cite{wang-etal-2020-efficient}&  INIT  & TC, QA &  Life-long learning \\
\cite{wu2021meta-lifelong-re}&  INIT   & IE &  Life-long learning \\
\cite{obamuyide-vlachos-2019-meta}&  INIT   & IE &  Life-long learning \\
     \hline
    \bottomrule
  \end{tabular}}
\end{table*}

\begin{table*}[h!]
  \centering
  \small
  \caption{Summary of learn-to-init variants. 
This table contains the following information. 
  (1)  \textbf{Method}: There are many variants in the learn-to-init family.
  The most representative one is MAML.
  Typical MAML~\cite{Finn:ICML17} has large computation intensity, so the first-order approximations like FOMAML~\cite{Finn:ICML17} and Reptile~\cite{nichol2018firstorder} are widely used. 
   DG-MAML~\citep{li2018metaDG} is for domain generalization.
  Typical learn-to-init assumes that all the tasks use the same network architecture, but LEOPARD~\citep{Bansal:arXiv19} and Proto(FO)MAML~\citep{Triantafillou2020Meta-Dataset} are proposed to overcome the limitation. 
  (2)  \textbf{How to Initialize the Initialization}: Learn-to-init approaches aim at learning the initial parameters. 
  But where does the initialization of MAML come from? 
  We found that using self-supervised pre-training as initialization is common. 
  The table specifies the pre-trained models used to initialize the learn-to-init methods. 
  '-' means the initial parameters are learned from random initialization or cannot tell based on the descriptions in the papers.
}
  \label{table:maml}
  \begin{tabular}{|l|l|l|}
    \toprule
Work  & Method & How to Initailize the Initailization \\
    \midrule
\citep{Bansal:arXiv19}  & LEOPARD & BERT   \\
\citep{Li_Wang_Yu_2020}  & MAML & Word Embedding   \\
\citep{unsupervisedMT-acl21}  & MAML &  XLM    \\ 
\citep{Gu:EMNLP18}  & FOMAML & Word Embedding   \\ %using universal multilingual representation, embedding is inited by  fasttext \\%enm+enc is the best
\citep{multiparsing}  & FOMAML & mBERT  \\ % + pre-train on English
\citep{chen-etal-2020-low}  & Reptile & BART   \\
\citep{Huang:ACL20}  & MAML & BERT  \\ %BERT is fixed (only for embedding)
\citep{wang-etal-2021-variance}  &  Propose a new method based on Reptile & Word Embedding    \\ %apply on several tasks

\citep{dingliwal-etal-2021-shot}  & Reptile & RoBERTa   \\ %finetune multiple times for mean \& STD \\
\citep{Qian:ACL19}  & MAML & Word Embedding  \\
\citep{ST-dialogue-AAAI21}  & MAML & Word Embedding   \\
\citep{Madotto:ACL19}  & MAML & Word Embedding   \\
\citep{dai-etal-2020-learning}  & MAML & -   \\
\citep{Hsu:ICASSP20}  & FOMAML & Multilingual ASR   \\
\citep{ASR-sample-AAAI21}  & MAML/FOMAML/Reptile & -   \\ %it is not clear it is pre-trained or not. 
\citep{Winata:INTERSPEECH2020}  & MAML & Pretrain by Supervised Learning   \\
\citep{Klejch:ASRU19}  & FOMAML & -   \\ %The last paragraph of Section 2.3 mentioned first order
\citep{SPmetaNLP21}  & MAML/FOMAML & -  \\ %FOMAML is better for without noise; MAML is better for with noise \\
\citep{Indurthi:arXiv19}  & FOMAML & -  \\
\citep{Winata:ACL2020} & FOMAML & -   \\
\citep{SP-ICASSP21}  & MAML & Pretrain by Multi-task Learning \\ %ANIL(encoder+decoder, or separator) & MAML is the best in most cases, but ANIL_s is better for libri+noise \\
\citep{ke-etal-2021-pre}  & MAML & BERT  \\ %special application \\
\citep{xia-etal-2021-metaxl}  & MetaXL & mBERT/XLM-R   \\ %NER / sentiment \\
\citep{Dou:EMNLP19}  & MAML/FOMAML/Reptile & BERT  \\
\citep{Obamuyide:ACL19}  & FOMAML & Word Embedding    \\
\citep{Lv:EMNLP19}  & MAML & -  \\ %using reinforce agent \\
\citep{holla-etal-2020-learning}  & FOMAML/Proto(FO)MAML & Word Embedding/ELMo/BERT \\ %EMNLP Finding
\citep{huang-etal-2020-towards-low} & MAML & Word Embedding \\ 
\citep{Mi:IJCAI19}  & MAML & -   \\ % It looks like the whole model is fined-tuned, but unclear in the paper \\ 
\citep{wang-etal-2021-meta}  & DG-MAML & BERT   \\ % MAML's variant, minimize support set and query set loss at the same time ... \\
%\citep{ranking-acl21}  & Meta-reweight & BERT   \\ %先meta learn不同筆data的weight (用類似MAML的bilevel optimization，inner update model，outer update weight)，再用學到的weight走一步正常的traiining
\citep{conklin2021meta}  & DG-MAML & -  \\
\citep{mhamdi-etal-2021-x}  & MAML & mBERT   \\ %分成meta train和meta adapt，meta train的support set是high-resource language，meta train的query和meta adapt的support&query都是low-resource
\citep{nooralahzadeh-etal-2020-zero}  & MAML & BERT/mBERT/XLM-R   \\ %selecting on or two auxilary langauges for MAML \\ 
%\citep{MetaKD-arXiv21}  & MAML & teacher yes, student no (so no ?) & whole & \\ %有兩個student，s1透過quiz examples update teacher to T', then T' distils to s2 %(European Conference on Artificial Intelligence)
%\citep{}&arXiv & 2020 & MAML(parameters are updated per batch-of-tasks instead of per task) & yes & whole & is this paper related to HLP? No HLP experiments \\
\citep{garcia-etal-2021-cross}  & MAML & mBERT    \\ %the meta-tasks is structure to a tree (relatedness between tasks), to MAML gradient propogates according to that tree \\
\citep{van-der-heijden-etal-2021-multilingual}   & FOMAML/Reptile/Proto(FO)MAML & XLM-R   \\
\citep{bansal-etal-2020-self}  & LEOPARD  &  BERT  \\ %SMLMT
\citep{murty-etal-2021-dreca}  & FOMAML &  BERT  \\
\citep{hua-etal-2020-shot}  & Reptile & -  \\     

 \cite{yan-etal-2020-multi-source} &  MAML  & BERT/RoBERTa \\
  \cite{wang2019tackling} & Reptile & - \\
  
      \cite{Bose:arXiv19} & Meta-Graph & - \\
      
     \hline
    \bottomrule
  \end{tabular}
\end{table*}

\begin{table*}[ht]
\caption{Performance of selected NAS approaches on the test set of GLUE benchmark.}
\label{table:NAS_Performance_on_GLUE}
\begin{center}
\small
%\begin{tabular}{ l|ll|llllllll|l } 
\resizebox{\textwidth}{11.5mm}{
\begin{tabular}{ l|ll|llllll|l } 
 
 \hline 
 %Model & \#Params & Latency & MNLI & QQP & QNLI & SST-2 & CoLA & STS-B & MRPC & RTE & Avg \\ \hline
 Model & \#Params & Latency & MNLI & QQP & QNLI & SST-2 & MRPC & RTE & Avg \\ \hline
 %BERT\textsubscript{BASE} (Google) & 108.9M & 362ms & 84.6 & 71.2 & 90.5 & 93.5 & 52.1 & 85.8 & 88.9 & 66.4 & 79.6\\
 BERT\textsubscript{BASE} (Google) & 108.9M & 362ms & \textbf{84.6} & \textbf{71.2} & 90.5 & \textbf{93.5} & \textbf{88.9} & 66.4 & \textbf{82.5}\\
 %MobileBERT\textsubscript{TINY} \citep{sun-etal-2020-mobilebert} & 15.1M & 96ms & 81.5 & 68.9 & 89.5 & 91.7 & 46.7 & 80.1 & 87.9 & 65.1 & 77.0\\ \hline
 MobileBERT\textsubscript{TINY} \citep{sun-etal-2020-mobilebert} & 15.1M & 96ms & 81.5 & 68.9 & 89.5 & 91.7 & 87.9 & 65.1 & 80.8\\ \hline

 AdaBERT \citep{ijcai2020-341} & 6.4-9.5M & 12.4-28.5ms & 81.6 & 70.7 & 86.8 & 91.8 & 85.1 & 64.4 & 80.1\\
 EfficientBERT \citep{dong-etal-2021-efficientbert-progressively} & 16M & 103ms & 83.0 & \textbf{71.2} & \textbf{90.6} & 92.3 & \textbf{88.9} & \textbf{67.8} & 82.3\\ \hline
 %EfficientBERT \citep{dong-etal-2021-efficientbert-progressively} & 16M & 103ms & 83.0 & \textbf{71.2} & \textbf{90.6} & 92.3 & 42.5 & 83.6 & 88.9 & 67.8 & 78.0\\ \hline

\hline 
\end{tabular}}
\end{center}
\end{table*}